\documentclass[11pt,a4paper]{article}
\usepackage[hyperref]{eacl2021}
\usepackage{times}
\usepackage{latexsym}
\usepackage{booktabs}
\usepackage{graphicx}
\usepackage{subcaption}
\usepackage{mwe}

\usepackage{microtype}

\aclfinalcopy % Uncomment this line for the final submission
\title{Identify, Align, and Integrate: Matching Knowledge Graphs to Commonsense Reasoning Tasks}

\author{Lisa Bauer \\
  UNC Chapel Hill\\
  \texttt{lbauer6@cs.unc.edu} \\\And
  Mohit Bansal \\
  UNC Chapel Hill\\
  \texttt{mbansal@cs.unc.edu} \\}

\date{}

\begin{document}
\maketitle
\begin{abstract}
Integrating external knowledge into commonsense reasoning tasks has shown progress in resolving some, but not all, knowledge gaps in these tasks. For knowledge integration to yield peak performance, it is critical to select a knowledge graph (KG) that is well-aligned with the given task's objective.  We present an approach to assess how well a candidate KG can correctly identify and accurately fill in gaps of reasoning for a task, which we call KG-to-task match. We show this KG-to-task match in 3 phases: knowledge-task identification, knowledge-task alignment, and knowledge-task integration. We also analyze our transformer-based KG-to-task models via commonsense probes to measure how much knowledge is captured in these models before and after KG integration. 
Empirically, we investigate KG matches for the SocialIQA (SIQA)~\cite{sap2019socialiqa}, Physical IQA (PIQA)~\cite{bisk2019piqa}, and MCScript2.0 ~\cite{ostermann2019mcscript2} datasets with 3 diverse KGs: ATOMIC~\cite{sap2019atomic}, ConceptNet~\cite{speer2017conceptnet}, and an automatically constructed instructional KG based on WikiHow~\cite{koupaee2018wikihow}.  With our methods we are able to demonstrate that ATOMIC, an event-inference focused KG, is the best match for SIQA and MCScript2.0, and that the taxonomic ConceptNet and  WikiHow-based KGs are the best matches for PIQA across all 3 analysis phases. We verify our methods and findings with human evaluation.\footnote{Code and commonsense probes: \url{https://github.com/lbauer6/IdentifyAlignIntegrate}}
\end{abstract}

\section{Introduction}
Recently, several datasets~\cite{sap2019socialiqa, huang2019cosmos, bhagavatula2019abductive, talmor2019commonsenseqa} have been released to tackle the challenge of commonsense reasoning. While deep pretrained language-models (LMs)~\cite{devlin2018bert, radford2019language, liu2019roberta, yang2019xlnet}  have been at the top of most leaderboards, they still have shortcomings when it comes to commonsense reasoning~\cite{sap2019socialiqa, rajani2019explain, mitra2019exploring}. Thus, incorporating knowledge graph (KG) information into these models is an active area of research~\cite{lin2019kagnet, sun2019ernie, mitra2019exploring, bosselut2019comet}. However, when selecting a KG match for a task, it is often difficult to quantitatively assess what kind of knowledge is missing from these models and how much of the missing knowledge required for the task is available in a candidate KG. It is also critical to examine how easily transformer-based models can learn commonsense knowledge, to determine the benefits of integrating a KG. 

\begin{figure*}[t]
	\centering
    \includegraphics[clip, width=0.82\textwidth]{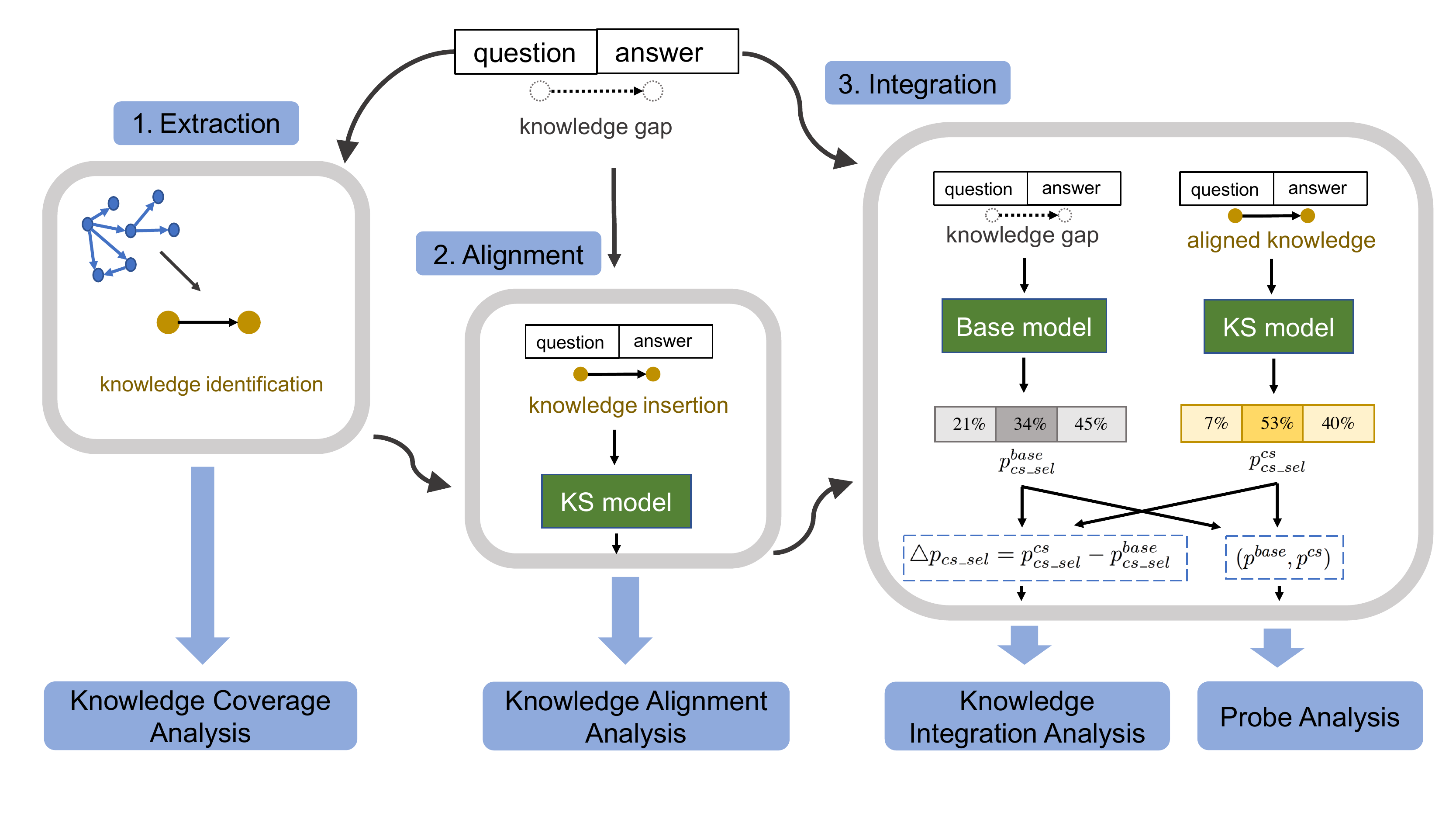}
     \caption{We illustrate our 3 phases of analysis: Knowledge-Task Extraction Analysis, Knowledge-Task Alignment Analysis, and Knowledge-Task Integration Analysis with Probing Analysis.  \label{fig:analysis_flow}} 
     \vspace{-9pt}
\end{figure*}

We investigate how well a KG matches with a task objective, referred to as \emph{KG-to-task match}. We use a 3-step process that examines knowledge \emph{identification, alignment,} and \emph{integration}. We utilize a modular pipeline approach to allow for interpretable results and easy replacement of new and different modules.
Our approach reveals features such as: how often a KG identifies a knowledge gap in a question-answer pair (identification), whether a KG identifies the correct knowledge gap (alignment), and whether the inserted knowledge correctly fills the knowledge gap required for the task (integration). These steps are depicted in Fig. \ref{fig:analysis_flow}. We also compare the effects of knowledge content, structure, and shape. 

The results of this analysis are impacted by the model we use, and thus we also develop \emph{probes} to examine how much commonsense knowledge LMs already know and how easy it is for them to learn. We evaluate our KG-to-task models in a QA probe setup to examine how much commonsense is learned with and without the matched KG. Our probes are automatically built from ATOMIC, enabling us to leverage existing knowledge sources as a probing base without relying on expensive collection methods. We also include an MLM probe setup to obtain zero-shot and fine-tuned results on probes for social relations, agent-patient assignment, and world knowledge.

We present detailed empirical results on three diverse datasets: SocialIQA (SIQA) task~\cite{sap2019socialiqa}, which requires social knowledge;  Physical IQA (PIQA)~\cite{bisk2019piqa} which requires physical knowledge; and MCScript2.0 ~\cite{ostermann2019mcscript2}, which requires commonsense script knowledge, not restricted to a particular domain.
Since both SIQA and PIQA require a particular domain of commonsense knowledge, these tasks allow us to draw strong conclusions about KG integration, as knowledge must be well aligned with the tasks to yield performance gains. Analyzing MCScript2.0, on the other hand, allows us to understand how this analysis applies to a task where the best match is not obvious. We compare KG-to-task match with three diverse KGs: ATOMIC~\cite{sap2019atomic}, ConceptNet~\cite{speer2017conceptnet}, and automatically extracted subgraphs from WikiHow. Each KG is tailored for a different commonsense domain: ATOMIC focuses on social commonsense, ConceptNet on taxonomic commonsense, and WikiHow on instruction-based commonsense. This allows us to see how different tasks require different types of commonsense knowledge.

To investigate KG-to-task match, we follow three phases: identify, align, and integrate. In our first phase, we examine knowledge gap identification by analyzing our extraction quantities. In our second phase, we examine alignment by utilizing a `knowledge-surrounded' (KS) model, in which we replace task candidate answers with knowledge-surrounded answers. We found that ATOMIC is the best match for SIQA across both identification and alignment: 11\% more ATOMIC data is extracted for question-answer knowledge gaps than ConceptNet data, with a 4.8\% performance increase over BERT using our ATOMIC KS model. We use our third phase, integration, to investigate the classification change distributions from BERT to the KS model, finding that our model is more confident about correct classification changes, supporting the ATOMIC-SIQA match. Additionally, both ConceptNet and WikiHow graphs outperformed ATOMIC on PIQA: 8\% more ConceptNet data is extracted than ATOMIC and a 17.4\% performance increase is achieved with our ConceptNet KS model, whereas we get a 15.5\% increase with our WikiHow KS model.
Finally, we find that ATOMIC is the best match for MCScript2.0, with a 2.7\%  increase with our ATOMIC KS model.

We also perform human evaluation and show important connections between the analysis phases. We see that if our KS model shows improvement for high quality settings, our extraction step is a valid knowledge-gap identification metric between 74\% and 89\% of the time, depending on the dataset. We also show that our best alignment strategy for ATOMIC-SIQA fills knowledge gaps 66\% of the time, outperforming the best alignment strategy for ConceptNet-SIQA, which supports our KS model performance results. We find similar trends for PIQA alignment and also find that the amount of information available at inference time may affect alignment results for MCScript2.0. Human evaluation shows that 93\% of ATOMIC-SIQA KS model prediction changes (with respect to the baseline) select the answer with the highest knowledge quality, verifying our integration phase as a quality metric.

Our commonsense QA probes before and after KG integration show that our KS model only considerably outperforms the BERT baseline on certain relational probes, indicating the type of knowledge gaps ATOMIC is better at resolving, e.g., relational knowledge such as feelings, reactions, etc.

Overall, our methods not only illustrate the type of knowledge that current transformer-based models are missing to approach human-level commonsense reasoning but also how we can identify, align, and integrate knowledge between a KG and a task to find the best match to fill in these missing gaps of reasoning.

\section{Related Work}
\noindent\textbf{Language Model Probes:} Recent work in probe construction has examined neural model knowledge~\cite{richardson2019does, zhou2019evaluating, rogers2020primer, lin2020birds}. \newcite{talmor2019olmpics} constructed eight tasks that evaluated LMs for operations such as comparison, conjunction, and composition. \newcite{zhou2020can} created logically equivalent probes to evaluate robustness on commonsense tasks to syntax. \newcite{kwon2019masked} proposed tests based on ConceptNet to measure what types of commonsense MLMs understand. Our work instead focuses on probing models for causal, social commonsense in both the MLM and QA setup before and after KG integration and fine-tuning, and automatically constructs probes from existing knowledge sources.\\
\noindent\textbf{Commonsense Reasoning:}
Recent commonsense reasoning datasets~\cite{bhagavatula2019abductive, zellers2018swag, zhou2019going, sap2019socialiqa, bisk2019piqa, lin2019comgen, zellers2019recognition, ostermann2019mcscript2} have motivated research in several domains of commonsense: abductive, grounded, temporal, social, and physical. Commonsense reasoning can be learned either by KGs pre-training ~\cite{bosselut2019comet, bosselut2019dynamic, ye2019align} or by integrating explicit knowledge~\cite{chen2017neural, mitra-etal-2018-incorporating, bauer2018commonsense, lin2019kagnet, zhang2019ernie, xiong2019improving}. We show how finding nuanced knowledge for successful commonsense reasoning can be quantitatively examined. \\
\noindent\textbf{Commonsense Knowledge Analysis:}\
\newcite{zhang2020winowhy} presented a categorization of essential knowledge for the Winograd Schema Challenge \cite{levesque2012winograd} via human annotation to identify what knowledge was required for better commonsense reasoning.  
\newcite{ma2019towards} investigated how KG integration methods affected model performance on different tasks and found that the degree of domain overlap between the KG and the task plays a crucial role in performance. We further investigate this by measuring KG-to-task match across 3 automatic phases, considering different extraction methods, and probing models for knowledge before and after KG integration.

\section{Tasks \& Knowledge Graphs}
\subsection{Tasks}
\textbf{SIQA}: The SocialIQA (SIQA)~\cite{sap2019socialiqa} task focuses on social commonsense. Given a context and question, a model selects from 3 answers. SIQA contexts are based on ATOMIC~\cite{sap2019atomic} events and SIQA question types are guided by ATOMIC inference dimensions. Thus, we expect ATOMIC to match SIQA requirements. For simplicity, we refer to the concatenation of context and question as the question throughout the paper. \\
\textbf{PIQA}:
The PhysicalIQA (PIQA)~\cite{bisk2019piqa} task objective focuses on physical commonsense reasoning. Given a goal, a model selects from 2 candidate solutions. PIQA is derived from the instruction domain, and thus we expect instructional physical commonsense to benefit PIQA. For simplicity, we refer to the goal as the question.\\
\textbf{MCScript2.0}: MCScript2.0 ~\cite{ostermann2019mcscript2} focuses on script events and participants, requiring commonsense knowledge, in particular script knowledge, to answer questions correctly. We specifically choose this dataset such that it does not have a strong preference for any of the KGs we investigate, to illustrate what our analysis may look like for an unpredictable result. For simplicity, we refer to the concatenation of context and question as the question throughout the paper.

\begin{figure*}[t]
	\centering
    \includegraphics[clip, height=150pt, width=0.85\textwidth]{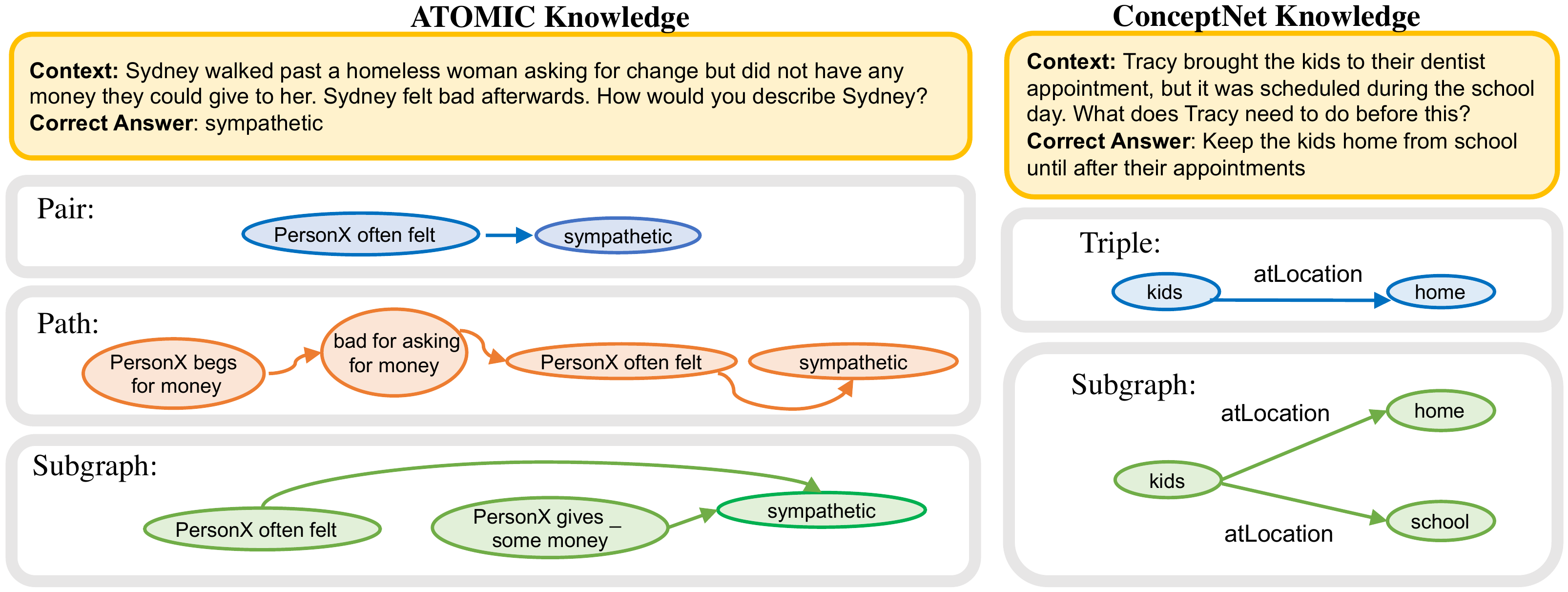}
     \caption{Examples of different knowledge shapes per KG given SIQA context and ground truth answer.  \label{fig:knowledge_shape}} 
\end{figure*} 

\subsection{Knowledge Sources}
We show results across three knowledge graphs to illustrate differences in KG-to-task identification, alignment, and integration, and to show how BERT responds differently to various types of knowledge. \\
\textbf{ATOMIC}: ATOMIC~\cite{sap2019atomic} is an inferential knowledge atlas that focuses on if-then reasoning. This knowledge is structured as event-inference pairs, where each pair reflects one of 9 possible inference dimensions for different if-then reasoning relations (cause vs. effect, etc). Knowledge is in the form of short, abstracted, free-text. See the left of Fig \ref{fig:knowledge_shape} for examples. \\
\textbf{ConceptNet}: ConceptNet~\cite{speer2017conceptnet} is a taxonomic knowledge graph that connects natural language concepts with relation edges. While there are relation edges similar to ATOMIC inference dimensions, the structure of the knowledge is in the form of structured triples and generally tends to focus on relations between words or phrases. See the right of Fig \ref{fig:knowledge_shape} for examples.\\
\textbf{WikiHow}: We automatically extract subgraphs from WikiHow~\cite{koupaee2018wikihow} to build our own instruction-based, domain-specific KG. Details are found in the appendix.

\begin{table}
    \centering
    \begin{small}
    \begin{tabular}{lccccc}
    \toprule
    \textbf{KG} & \textbf{Shape} & \textbf{Cond.} & \textbf{Filter} & \textbf{Sets} & \textbf{Pres.}\\
    \midrule
    AT & triple& QC& HQ&CS-1&KS+\\
    CN & path & A & HR&CS-2&KS-\\ 
    WH&subgraph&-&-&CS-3&-\\ 
    \bottomrule
    \end{tabular}
    \end{small}
    \caption{Variations for each knowledge setting. AT=ATOMIC, CN=ConceptNet, WH=WikiHow, QC=question-conditioned, A=unconditioned, HR=high recall, HQ=high quality, KS+/-=knowledge presence at inference time.}
    \vspace{-10pt}
    \label{tab:setup}
\end{table}

\section{Phase 1: Identify}
\subsection{Setup}
\label{sec:identifysetup}
We identify knowledge using the following extraction methods for each KG. Our setup with all possible options is illustrated in Table~\ref{tab:setup}. We will use Fig \ref{fig:knowledge_shape} as a running example throughout this section.

\subsubsection{Knowledge Conditioning}
\textbf{Unconditioned (A) Answer-Knowledge:}\\
\textit{ATOMIC}: For each candidate answer, we extract a pool of top scoring knowledge using tf-idf between the answer and all ATOMIC event-inference pairs. \\
\textit{ConceptNet \& WikiHow}: For each candidate answer, we extract knowledge that links concepts in the answer to any concept in the KG, where concepts are tokens in the answer and nodes in the KG. \textit{Example}: Consider the SIQA context and ground-truth answer on the right side of Fig \ref{fig:knowledge_shape}. Here, the  A conditioning setup for ConceptNet would extract the triple [keep, Antonym, get\_rid]. 

\noindent\textbf{Question-Conditioned (QC) Answer-Knowl.:}\\
\textit{ATOMIC}: We select a question-conditioned knowledge pool via the top scoring tf-idf match between the question \& candidate answer and all ATOMIC event-inference pairs. We then select a pool of top scoring knowledge for each candidate answer using tf-idf between the candidate answer and the question-conditioned knowledge pool. \\
\textit{ConceptNet \& WikiHow}: For each candidate answer, we extract knowledge that links concepts in the question directly to concepts in the answer. \\
\textit{Example}: All knowledge illustrated in Fig \ref{fig:knowledge_shape} is extracted using QC conditioning.

\subsubsection{Knowledge Shape}
\noindent\textbf{Knowledge Pairs/Triples:}\\
\textit{ATOMIC}: We take the highest scoring knowledge pair determined by the conditioning step. \\
\textit{ConceptNet \& WikiHow}: We select a triple at random from the conditioning step. \\
\textbf{Knowledge Paths:} \\
\textit{ATOMIC}: In the QC setup, for each data point, we extract a question-knowledge pool via top scoring tf-idf match between the question and all ATOMIC event-inference pairs. If there exists a concept link between the question-knowledge pool and the answer-knowledge pool from the conditioning step,  we link this knowledge as a path. In the A setup, we make the modification that our answer-knowledge pool can link to any pair in ATOMIC. \\
\textit{WikiHow}: In the QC setup, we find a path from a word in the question, to another word in the question, to a word in the answer. In the A setup, we find a path from a word in the answer to any word it connects to in the KG, as a path through the KG.
\\
\textbf{Knowledge Subgraphs:}\\
\textit{ATOMIC}:  We take a maximum of the 3 highest scoring knowledge triples determined by the conditioning step to create 1-hop subgraphs. \\
\textit{ConceptNet \& WikiHow}: From the conditioning knowledge pool, we add the subgraph with the highest number of edges, as we assume these to be the most informative. We only consider 1-hop edges and take the top 5. \\
\textbf{Example}: All three shape variations are illustrated in Fig \ref{fig:knowledge_shape}, using QC conditioning.

\subsubsection{Knowledge Filtering}
\noindent\textbf{High Quality/Low Recall (HQ):} We constrain each answer candidate to keep its highest scoring unique knowledge such that no answer candidate shares knowledge, intending to ensure the relevance of the knowledge to that candidate alone. \\
\noindent\textbf{Low Quality/High Recall (HR):} Candidate keeps its highest scoring knowledge regardless of knowledge sharing among candidates. \\

\subsubsection{Data Subsets \& Baseline Training}
We split data into subsets depending on how many candidate answers extracted knowledge (CS-X) to evaluate knowledge impact on task performance fairly. For our main results, we use the split in which each answer has access to knowledge (CS-2 for PIQA and MCScript2.0, CS-3 for SIQA). Table \ref{tab:data_stats} illustrates the percent of original data for each split. We compare KS model subset results against a BERT baseline trained and evaluated on the same subset simply without the added knowledge.

\subsection{Analysis}
\begin{table}[t]
    \centering
    \begin{small}
    \begin{tabular}{lcc}
    \toprule
    \textbf{Variation} & \textbf{ATOMIC} & \textbf{ConceptNet}\\
    \midrule
    SIQA: &&\\
    QC-HQ CS-1 & \textbf{51}\% &  39\%\\ 
    QC-HQ CS-2 & \textbf{24}\%& 22\%\\
    QC-HQ CS-3  & 2\% & \textbf{6}\%\\
    QC-HQ CS & \textbf{77}\% & 66\%\\
    \midrule
    PIQA: &&\\
    QC-HQ CS-1 & 16\% & \textbf{24}\%\\ 
    QC-HQ CS-2 & 8\%& 8\%\\
    QC-HQ CS & 24\% & \textbf{32}\%\\
    \midrule
    MCScript2.0: &&\\
    QC-HQ CS-1 & 2\% & \textbf{36}\%\\ 
    QC-HQ CS-2 & \textbf{88}\%& 54\%\\
    QC-HQ CS & 90\% & 90\%\\
    \bottomrule
    \end{tabular}
    \end{small}
    \caption{\%Knowledge extracted for each subset wrt. original data size. Results shown for the best aligned KG shape in the QC-HQ setting: SIQA=CN triples, ATOMIC paths; PIQA=CN subgraphs, ATOMIC pairs; MCScript2.0=CN subgraphs, ATOMIC pairs.}
    \vspace{-10pt}
    \label{tab:data_stats}

\end{table}
We examine how often a KG identifies a potential knowledge gap between a question and an answer. This is illustrated on the far left in Fig. \ref{fig:analysis_flow}.
Table \ref{tab:data_stats} shows the percent of knowledge extracted for each QC-HQ subset. We use the QC-HQ setting to show how often our KG specifically identifies question-answer knowledge gaps. We use each KG’s best aligned shape in this comparison (Section \ref{sec:knowledge_shape}). For SIQA, we extract more ATOMIC data than for ConceptNet and for PIQA, we extract more ConceptNet data than for ATOMIC. This illustrates that ATOMIC identifies more knowledge gaps for SIQA, and ConceptNet identifies more knowledge gaps for PIQA. For MCScript2.0, we see that the same total knowledge is extracted from both KGs, however more ATOMIC knowledge is extracted in the CS-2 setup, indicating better coverage.

\section{Phase 2: Align}
\subsection{Setup}
\noindent \textbf{Baseline Model:}
We fine-tune BERT-base~\cite{devlin2018bert} as our baseline on SIQA following \newcite{sap2019socialiqa}, BERT-base on MCScript2.0, and BERT-large~\cite{devlin2018bert} as our baseline on PIQA following \newcite{koupaee2018wikihow}. See original papers for hyperparameter settings.

\noindent \textbf{Knowledge-Surrounded (KS) Model:}
We enhance task candidate answers with knowledge-surrounded answers, in which answer-specific knowledge is appended to each candidate answer. This knowledge is intended to explicitly add missing knowledge gaps to the answer. This model is used in the \textit{alignment} and \textit{integration} steps in Fig. \ref{fig:analysis_flow}. 
We encode input as follows. For a question $q_i$, we modify each candidate answer, $c_{ij}$, by appending its respective knowledge $k_{ij}$. The sequence of tokens \{[CLS] $q_i$ [SEP] $c_{ij}k_{ij}$ [SEP]\} is then passed as input to BERT. Thus, each candidate answer is surrounded by knowledge that allows BERT to potentially fill reasoning gaps between the question and answer. The extraction variations for  $k_{ij}$ are described in Section \ref{sec:identifysetup}.

\subsection{Analysis}
We investigate how well the extracted knowledge and the task are aligned by allowing the knowledge to fill in the question-answer knowledge gap and determining whether this improves performance. This is illustrated in the center of Fig. \ref{fig:analysis_flow}.
\begin{table}[t]
    \centering
    \begin{small}
    \begin{tabular}{l|ccc}
    \hline
    \textbf{Variation} & \textbf{Base.} & \textbf{KS+}  & \textbf{KS-}\\
    \hline
    SIQA AT. QC-HQ& 38.1 & \textbf{42.9} & 40.5 \\
    SIQA AT. QC-HR & \textbf{60.3} & 59.0 & 56.4\\
    SIQA AT. A-HQ & \textbf{60.4} & 60.0 & 59.2 \\
    SIQA AT. A-HR & \textbf{61.8} & 60.8 & 61.0 \\
    \hline
    SIQA CN QC-HQ  & \textbf{54.1} & 45.0& 43.2 \\
    SIQA CN QC-HR & \textbf{54.5} & 52.2 & 49.1\\
    SIQA CN A-HQ  & 61.5 & 61.5& \textbf{61.9}\\
    SIQA CN A-HR& \textbf{61.2} & 60.0 & 59.7\\
    \hline
    PIQA AT. QC-HQ & \textbf{54.6} & 49.3 & 50.0\\
    PIQA AT. QC-HR & \textbf{61.8} & 51.2 & 48.8\\
    PIQA AT. A-HQ & 50.6& 61.2 & \textbf{64.9}\\
    PIQA AT. A-HR & \textbf{70.5} & 64.9 & 65.4\\
    \hline
    PIQA CN QC-HQ & 60.5 & \textbf{66.5} & 59.9\\
    PIQA CN QC-HR & 51.4 & 57.2 & \textbf{58.7}\\
    PIQA CN A-HQ &49.0 & 64.5 & \textbf{66.4}\\
    PIQA CN A-HR  & \textbf{70.5} & 54.5 & 51.9\\
    \hline
    PIQA WH QC-HQ & 53.3  & \textbf{54.7} & 48.0\\
    PIQA WH QC-HR &\textbf{59.9} & 54.7&  55.0\\
    PIQA WH A-HQ & 53.5 & 68.3 & \textbf{69.0}\\
    PIQA WH A-HR & \textbf{67.5} & 51.4 & 49.3\\
    \hline
    MC AT. QC-HQ & 80.3 &  \textbf{83.0} & 79.6\\
    MC AT. QC-HR & \textbf{82.4} & 80.8 & 79.4\\
    MC AT. A-HQ & \textbf{82.5} & 80.9 & 79.6\\
    MC AT. A-HR & \textbf{81.2} & 80.0 & 77.6\\
    \hline
    MC CN QC-HQ & 78.6 & \textbf{79.4} & 76.2\\
    MC CN QC-HR & 78.8 & \textbf{79.3} & 78.4\\
    MC CN A-HQ &\textbf{82.6} & 80.3 & 77.4\\
    MC CN A-HR  &\textbf{82.2} & 80.2  & 76.9\\
    \hline
    \end{tabular}
    \end{small}
    \vspace{-4pt}
    \caption{Accuracy for extraction variations on: SIQA CS-3 for ATOMIC paths \& ConceptNet triples;  PIQA CS-2 for ATOMIC pairs \& ConceptNet subgraphs \& WikiHow paths; and MCScript2.0 CS-2 for ATOMIC pairs \& ConceptNet subgraphs.} 
    \vspace{-10pt}
    \label{tab:main_results}

\end{table}

\subsubsection{Extraction Variation Analysis}
Table \ref{tab:main_results} illustrates performance for each KG across each of the different extractions, using the subset in which each candidate answer has access to knowledge. We compare question-conditioned, unconditioned, high recall, and high quality settings for the best aligned knowledge shape (ConceptNet triples, ATOMIC paths for SIQA; ConceptNet subgraphs, ATOMIC pairs, WikiHow paths for PIQA; ConceptNet subgraphs, ATOMIC pairs for MCScript2.0) across the three KGs. For each setting, we show results both for when knowledge is present during inference time (KS+) and when it is not (KS-). We see that SIQA performed best when it received QC-HQ knowledge from ATOMIC, reflecting the strong, one-to-one alignment between SIQA and ATOMIC. PIQA, however, performs well across most extractions for ConceptNet, indicating that PIQA is generally well aligned with ConceptNet and only performs poorly when the extraction process becomes too noisy. Additionally, PIQA performs well with WikiHow for the unconditioned, high quality setting, indicating that the WikiHow KG is not well aligned across question-answer pairs, but does identify useful knowledge gaps within the answer that may improve performance on the task. Finally, we see that MCScript2.0 performed best when it received QC-HQ knowledge from ATOMIC, and similarly to SIQA, improves when seeing this knowledge at inference time.

\subsubsection{Knowledge Shape Analysis}
\label{sec:knowledge_shape}

We discuss knowledge shape effects on alignment. \\
\noindent\textbf{ATOMIC:}
For SIQA, ATOMIC paths have the best alignment, due to the high quality achieved when constraining knowledge for the SIQA question to link to knowledge for the answer. ATOMIC pairs and subgraphs seem to be learned more implicitly and do not yield large overall improvements when added explicitly during inference time. It seems that SIQA requires longer, more informative knowledge at inference time, which pairs and subgraphs do not offer.
For example, consider the SIQA context and answer on the right of Fig \ref{fig:knowledge_shape}. For this data point, we extract the following ATOMIC path: [PersonX has to go to the dentist, need to make an appointment, PersonX picks \_ up from school, to drive kids home], and the following ATOMIC pair: [PersonX picks \_ up from school, to drive kids home]. We can see that the path clearly contains more context and detail for the knowledge required to make the correct prediction.
For PIQA, we saw the largest improvements for ATOMIC pairs and subgraphs, where pairs ultimately perform best, indicating that PIQA might find concise and direct information from ATOMIC more useful. 
For MCScript2.0, ATOMIC pairs aligned best exclusively. \\
\noindent\textbf{ConceptNet:}
For SIQA, ConceptNet triples and subgraphs show similar alignment results and we do not see major improvements. It seems that the content of ConceptNet is not aligned well to SIQA, regardless of shape. 
For PIQA, we see improvements for both triples and subgraphs, and get our best improvements with subgraphs, indicating that the extra knowledge encoded in a subgraph shape via ConceptNet is helpful for the PIQA task.
Similarly to SIQA, MCScript2.0 performed best with ConceptNet subgraphs, but these do not yield major improvements. Interestingly, results on MCScript2.0 only show slight improvement for ConceptNet when knowledge is present at inference time, and shows no improvement otherwise.\\ 
\noindent\textbf{WikiHow:}
WikiHow paths performed best, indicating that paths were the best way to extract information, as WikiHow pairs and subgraphs might have contained redundant information given limitations with the WikiHow KG extraction process. 

\subsubsection{Knowledge Graph Analysis}
Given our alignment results, it is clear to see that ATOMIC is the best match for SIQA, that ConceptNet is the best match for PIQA, and that ATOMIC is the best match for MCScript2.0 (most likely due to its need for script knowledge, which often requires social knowledge).
The encoding of each KG plays an important role in this match. We see that the ConceptNet to PIQA match is more robust to extraction methods, which may be a side effect of ConceptNet's encoding, where directly linking nodes is less noisy than using tf-idf measures for the ATOMIC encoding, in which we only see positive results when we have very selective filters in our extraction techniques. The concise, short nature of ConceptNet's knowledge also lends itself to more implicit knowledge learning for certain types of tasks, whereas the more descriptive nature of ATOMIC can be read during inference time (see Fig \ref{fig:knowledge_shape} for examples). This illustrates the possibility that ConceptNet may boost performance as a regularizer for certain tasks. 

\begin{table}[t]
    \centering
    \begin{small}
    \begin{tabular}{l|cc}
    \toprule
    \textbf{Type} & \textbf{Baseline} & \textbf{KS model} \\
    \midrule
    \textbf{Relation Probes} &&\\
    xWant vs xEffect & \textbf{53.66} & {46.59}\\
    xWant vs xReact & 51.66 & \textbf{77.39}\\
    xWant vs xIntent & \textbf{52.05} & {45.77}\\
    xWant vs xNeed &  {47.99}  &  \textbf{50.32}\\
    xWant vs xAttr & \textbf{71.74} & 65.19\\
    xEffect vs xReact & 43.32 & \textbf{63.36}\\
    xEffect vs xIntent &  \textbf{54.29} & {44.85}\\
    xEffect vs xNeed & {49.22} & \textbf{52.48}\\
    xEffect vs xAttr &  \textbf{61.52} &  {59.44}\\
    xReact vs xIntent & 52.34 & \textbf{73.32}\\
    xReact vs xNeed & 49.21 & \textbf{74.76}\\
    xReact vs xAttr & 48.48 & \textbf{57.71}\\
    xIntent vs xNeed &  \textbf{44.18} & {42.24}\\
    xIntent vs xAttr & \textbf{78.96} & 64.64\\
    xNeed vs xAttr &  \textbf{68.97} &  {68.49}\\
    oWant vs oEffect &  \textbf{51.74} &  {41.83}\\
    oWant vs oReact & 43.90 & \textbf{72.04}\\
    oEffect vs oReact & 49.12 & \textbf{59.02}\\
    \midrule
    \textbf{Agent-Patient Probes}&&\\
    xWant vs oWant &  \textbf{70.12} &  {70.09}\\
    xEffect vs oEffect &  \textbf{74.34} &  {73.51}\\
    xReact vs oReact & \textbf{74.34} & 69.51\\
    \midrule
    \textbf{Concept Probes} &&\\
    inference & 74.96 & \textbf{77.40}\\
    event & \textbf{67.74} & 65.07\\
    \bottomrule
    \end{tabular}
    \end{small}
    \vspace{-5pt}
    \caption{QA probe results on ATOMIC-SIQA QC-HQ KS model vs BERT baseline.}
    \label{tab:main_results_qa_probes}
    \vspace{-5pt}
\end{table}

\section{Phase 3: Integrate}
\subsection{Setup}
We analyze two aspects of integration, depicted on the far right in Fig. \ref{fig:analysis_flow}. First, we construct commonsense probes to demonstrate how much knowledge we gain from our KGs via our transformer-based KS model with respect to a BERT baseline. Second, we examine distributional changes in our models before and after commonsense integration and verify our results with human evaluation.
With our probes, we can compare how well models distinguish between several types of ATOMIC-style knowledge, outlined below.\\
\noindent\textbf{Relational Probes:} We predict the ATOMIC relation between an event and inference pair, constraining our candidate answer set to two specified inference dimensions. For example, \texttt{xWant} vs \texttt{xNeed} might refer to a probe that will predict an answer from the candidate set: [Person \texttt{wants} recognition, Person \texttt{needs} recognition] given some event, essentially pitting the two relations against each other to evaluate the difficulty of distinguishing. \\
\noindent\textbf{Agent-Patient Probes:} We predict the agent of the inference where the candidate set is the agent and patient of the event (using ATOMIC abstractions). \\
\noindent\textbf{Concept Probes:} We predict concepts and constrain our candidate answer set to the most salient concept in the sequence and its respective antonym.\\ 
A full description of probe construction and examples for each knowledge type can be found in the appendix. We evaluate QA probes via standard accuracy after fine-tuning.  

\subsection{Analysis: Commonsense Probes}
Table \ref{tab:main_results_qa_probes}  compares the performance between the best performing ATOMIC-SIQA KS model and respective SIQA BERT baseline. We see notable performance differences for our relation probes, showing that the KS model does well at identifying the \texttt{React} relation and that the baseline tends to identify the \texttt{Attribute} relation well. While the KS model improves on several QA relational probes, the performance was comparable for agent-patient and world knowledge probes, indicating the knowledge gaps ATOMIC is better at resolving, i.e., relational (feelings, reactions, etc.) versus antonym/synonym information about concepts.

\subsection{Analysis: Distribution Change} 
We conduct an integration analysis on our best ATOMIC-SIQA setting (QC-HQ). 
We examine 40 multiple choice questions and analyze KS model prediction changes with respect to the baseline. We observe that 93\% of prediction changes were made because the new prediction's knowledge had the best reasoning flow to resolve a knowledge gap.

Fig. \ref{fig:analysis_flow} defines our distribution change analysis as $\Delta p_{cs\_sel} = p^{cs}_{cs\_sel} - p^{base}_{cs\_sel} $. Here, $p^{cs}_{cs\_sel}$ indicates the KS Model's probability of selecting the KS Model's selected answer and  $p^{base}_{cs\_sel}$ indicates the baseline's probability of selecting the KS Model's selected answer. Thus, $\Delta p_{cs\_sel}$ indicates the change in the probability of selection for the KS Model's selected answer before (Base Model) and after (KS Model) knowledge  integration.

Table \ref{tab:misclass_analysis} shows the distribution change from the baseline to the KS model for the selected answer. When a switch became positive, the average probability increase of the selected ground truth candidate answer was 19.5\%, whereas when a switch became negative the increase was 12.4\%. Thus, the distribution change shows more confidence about ground truth selection with added knowledge, indicating that the quality of a ground truth’s knowledge is higher than that of a negative candidate.

\subsection{Analysis: Human Evaluation} \label{sec:human_eval}

\begin{table}[t]
    \centering
    \begin{small}
    \begin{tabular}{lcc}
    \toprule
    \textbf{Correct} & \textbf{Incorrect}\\
    \midrule
    19.5\% & 12.4\%\\
    \bottomrule
    \end{tabular}
    \end{small}
    \caption{Distribution changes for selected class from baseline to KS model.}
    \label{tab:misclass_analysis}
    \vspace{-10pt}
\end{table}

We performed human evaluation on 100 SIQA, 100 PIQA, and 100 MCScript2.0 question-answer pairs to determine the validity of our process for both knowledge gap identification and alignment.\footnote{Done by expert authors since this is time-consuming, fine-grained verification analysis (instead of model evaluation).}
To show the validity of our QC-HQ extraction method as a measure for knowledge gap identification, we find that this extraction method is a valid potential SIQA knowledge gap identification 89\% of the time for ATOMIC and 91\% for ConceptNet. Valid, in this case, means that the correct concepts (that identify a relevant knowledge gap) were used to create a link. These results are found in Table \ref{tab:human_eval_valid}. 
We also show that for our best ATOMIC extraction (QC-HQ), we extract the correct knowledge for the gap 66\% of the time, demonstrating the connection between KS model improvement and alignment. Correct, in this case, means that the content of the link itself is relevant to resolve the commonsense gap. These results are found in Table \ref{tab:human_eval_correct}. In contrast, we see that our best ConceptNet extraction (A-HQ) finds the correct knowledge for the gap 18\% of the time. This is probably why we do not see much improvement when we give our ConceptNet KS model knowledge during inference time and why it seems to improve mostly via regularization. \\
On PIQA, we find that this extraction method is a valid potential knowledge gap  identification 48\% of the time for ATOMIC and 82\% for ConceptNet. We conclude that if we do not see alignment improvement on the QC-HQ setting (as is true of ATOMIC-PIQA), then extraction does not indicate the best knowledge gap coverage. Additionally, we find that for our best ATOMIC extraction method (A-HQ), we extract the correct knowledge for the gap 16\% of the time and that for our best ConceptNet extraction method (A-HQ), we extract the correct knowledge for the gap 22\% of the time. 

\begin{table}[t]
    \centering
    \begin{small}
    \begin{tabular}{lcc}
    \toprule
     & \textbf{ATOMIC} & \textbf{CN}\\
    \midrule
    SIQA & 89\% & \textbf{91}\%\\
    PIQA & 48\% & \textbf{82}\%\\
    MC & \textbf{75}\% & 74\%\\
    \bottomrule
    \end{tabular}
    \end{small}
    \vspace{-5pt}
    \caption{Human evaluation for valid knowledge gap.} 
    \label{tab:human_eval_valid}
    \vspace{-5pt}
\end{table}

\begin{table}[t]
    \centering
    \begin{small}
    \begin{tabular}{lcc}
    \toprule
     & \textbf{ATOMIC} & \textbf{CN}\\
    \midrule
    SIQA & \textbf{66}\% & 18\%\\
    PIQA & 16\% & \textbf{22}\%\\
    MC & 31\% & \textbf{43}\%\\
    \bottomrule
    \end{tabular}
    \end{small}
    \vspace{-5pt}
    \caption{Human evaluation for correct knowledge gap.} 
    \label{tab:human_eval_correct}
    \vspace{-10pt}
\end{table}

For MCScript2.0, we found the best empirical performance with QC-HQ settings for both ATOMIC and ConceptNet. With these settings, we found that a valid potential knowledge gap identification occurs 75\% of the time for ATOMIC and 74\% for ConceptNet.
Additionally, we find that with ATOMIC, we extract the correct knowledge for the gap for 31\% of examples, and with ConceptNet for 43\%. The higher correct extractions for ConceptNet are most likely due to the best performing extraction settings being QC-HQ. ATOMIC QC-HQ settings visibly outperform the baseline empirically, whereas ConceptNet QC-HQ settings perform only slightly better. This may be due to the fact that MCScript2.0 has a much larger context than any of our other datasets, and thus a model may already be able to implicitly infer the explicit taxonomic knowledge offered by ConceptNet.
\begin{table}[t]
    \centering
    \begin{small}
    \begin{tabular}{lccc}
    \toprule
    \textbf{Type} & \textbf{BERT}  & \textbf{RoBERTa} \\
    \midrule
    \textbf{Relation Probes} &&\\
    xWant vs xEffect & \textbf{60.90} &  56.68  \\
    xWant vs xReact & 94.15 & \textbf{94.96}  \\
    xWant vs xIntent&	57.02 & \textbf{57.10}  \\
    xWant vs xNeed 	& 57.13 &	\textbf{59.52}\\
    xWant vs xAttr &	\textbf{87.73}&		86.68	 \\
    xEffect vs xReact &	\textbf{60.88} &57.93 \\
    xEffect vs xIntent &	\textbf{78.74}&71.79\\
    xEffect vs xNeed &	\textbf{58.38}&	50.49\\
    xEffect vs xAttr	&\textbf{58.18}&	58.13\\
    xReact vs xIntent &	84.97&	\textbf{92.59} \\
    xReact vs xNeed 	&\textbf{94.63}&	94.29\\
    xReact vs xAttr	&56.72&	\textbf{57.27}\\
    xIntent vs xNeed &	51.23&	\textbf{52.67}\\
    xIntent vs xAttr &	61.57	&	\textbf{64.73}\\
    xNeed vs xAttr &	\textbf{83.51}	&	77.87	\\
    oWant vs oEffect&	\textbf{61.71}&		61.61\\
    oWant vs oReact &	\textbf{94.80}&90.97\\
    oEffect vs oReact &	60.46&	\textbf{61.00}\\
    \midrule
    \textbf{Agent-Patient Probes}&&\\
    xWant vs oWant&	\textbf{65.79}&	46.71\\
    xEffect vs oEffect &	32.24&	\textbf{60.67}\\
    xReact vs oReact &	\textbf{62.64}	&52.10\\
    \midrule
    \textbf{Concept Probes} &&\\
    inference & 79.15&	\textbf{80.68}\\
    event &\textbf{68.91}&	68.51 \\
    \bottomrule
    \end{tabular}
    \end{small}
    \caption{MLM probe zero-shot results.}
    \label{tab:mlm_probes}
\end{table}
\section{MLM Commonsense Probes} 
We evaluate the transformer-based models used in our setup to assess how much knowledge LMs already know and how easy it is for them to learn. 
\subsection{Setup}
We examine our MLM probes in two settings: zero-shot and fine-tuned. For the zero-shot setting, we use a pre-trained LM without any fine-tuning. This is to examine how much knowledge a pre-trained transformer model already holds. For the fine-tuned setting, we train on each probe's respective train set and evaluate using the same metrics as in \newcite{talmor2019olmpics}. This is to examine how fast a model learns given its encoding before fine-tuning. Results, set up, metrics, and analysis for fine-tuned settings are found in the appendix.

\subsection{Results}
Table \ref{tab:mlm_probes}  compares the performance of BERT and RoBERTa for zero-shot results. Majority label results are found in the appendix. 

\noindent\textbf{Zero-shot Results:} 
RoBERTa and BERT perform comparably for most Relation probes. While performance is poor for most settings, both models perform very well at discerning between \texttt{Want} and \texttt{React} (\texttt{xWant} vs \texttt{xReact}, \texttt{oWant} vs \texttt{oReact}), and between \texttt{xReact} vs \texttt{xNeed}.
Both models perform reasonably well at discerning \texttt{Attr} and \texttt{Intent} from other dimensions in certain settings (\texttt{xWant} vs \texttt{xAttr}, \texttt{xNeed} vs \texttt{xAttr}, \texttt{xEffect} vs \texttt{xIntent}, \texttt{xReact} vs \texttt{xIntent}). In general, the models seem to most consistently discern \texttt{React} from other dimensions. Finally, both models perform comparably and reasonably well on Concept probes, whereas the performance for Agent-Patient probes differs largely between the models and is often poor. 

\section{Conclusion}
We proposed a method to analyze how well a candidate KG can correctly identify and accurately fill in gaps of reasoning for a given task. 
We presented a three step approach for analyzing this KG-to-task match via identification, alignment, and integration. We found that the ATOMIC KG aligns best with the SIQA task, and quantitatively analyze the quality of the extracted commonsense. We also found that the ConceptNet and WikiHow based KGs match best with the PIQA task. Finally, we see that the ATOMIC KG also aligns best with MCScript2.0, which was a novel discovery and is most likely a result of the task's script knowledge requirement. We demonstrate the knowledge contained and learned by our KS model via our commonsense probes, illustrating what knowledge transformer-based models already know and what they can learn. This analysis can be extended to any set of tasks and KGs to analyze match potential. 

\section*{Acknowledgments}
We thank the reviewers for their useful feedback. This work was supported by DARPA MCS Grant \#N66001-19-2-4031, NSF-CAREER Award
1846185, NSF PhD Fellowship, and awards from Microsoft and Amazon. The views are those of the authors and not of the funding agency.

\bibliography{anthology,eacl2021}
\bibliographystyle{acl_natbib}

\appendix
\section{Appendix}

\subsection{WikiHow Subgraph Extraction Procedure}
We extract PIQA-conditioned WikiHow subgraphs for each PIQA datapoint. We do this in three steps:

\noindent 1. Given a PIQA goal, extract relevant titles in Wikihow via tf-idf. \\
2. Dependency parse the PIQA goal, extracted Wikihow title, and each sentence in the title's corresponding paragraph. \\
3. Find overlapping concepts in the dependency parses for which to create concept nodes. Then, create a graph by combining all possible edges for a concept node (found in all dependency parses).

\subsection{Probe Construction Details}
\begin{table}
    \centering
    \begin{small}
    \begin{tabular}{l|c}
    \toprule
    \textbf{Type} & \textbf{Data Size}\\
    \midrule
    xWant vs xEffect & 10012\\
    xWant vs xReact & 10693 \\
    xWant vs xIntent & 9790 \\
    xWant vs xNeed & 9829\\
    xWant vs xAttr & 11730\\
    xEffect vs xReact &  9519\\
    xEffect vs xIntent &8616 \\
    xEffect vs xNeed & 	8655\\
    xEffect vs xAttr &10556\\
    xReact vs xIntent &  9297\\
    xReact vs xNeed & 9336\\
    xReact vs xAttr & 11237\\
    xIntent vs xNeed & 8433\\
    xIntent vs xAttr &10334 \\
    xNeed vs xAttr & 10334\\
    oWant vs oEffect &3873\\
    oWant vs oReact &3873\\
    oEffect vs oReact  & 3685\\
    xWant vs oWant & 7974\\
    xEffect vs oEffect & 5911\\
    xReact vs oReact & 7293\\
    inference & 8401\\
    event & 13228\\
    \bottomrule
    \end{tabular}
    \end{small}
    \caption{MLM \& QA dev probe sizes.}
    \label{tab:probe_size}
\end{table}
\subsubsection{QA}
We use ATOMIC events and respective inferences and map them into the following QA formats.

\paragraph{Relational Probes.}
For relational probes, we state the event and follow it with ``What happens next?". We then create candidates out of a corresponding inference, each with a different relation. For example, 

\begin{center} PersonX puts out a fire. What happens next? \end{center}

For the probe \texttt{xWant} vs  \texttt{xNeed}, we select from the candidate answers: \textit{[PersonX wants to receive recognition, PersonX needs to receive recognition]}, where our ground truth is \textit{PersonX wants to receive recognition}.

\paragraph{Agent-Patient Probes.}
ATOMIC inference dimensions are either assigned to ``PersonX" (most often the agent of the event, unless otherwise specified) or ``others" (who are often influenced by the effects of PersonX's actions and may not be directly referred to in the event, see original ATOMIC paper for more details). 
For Agent-Patient probes, we state the event and follow it with ``Who [relation] [inference]"? 
Using our previous example, we have the following:
\begin{center} PersonX puts out a fire. Who wants to receive recognition? \end{center}

Where our candidate answers are: \textit{[PersonX, others]}. 

\paragraph{Concept Probes.}
For event concept prediction, we first state the inference and then ask ``What happened?" We then create event candidates with a ground truth salient concept (determined in the same way as the MLM salient concepts described in the next section) and an antonym concept.  For example:

\begin{center} PersonX wants to receive recognition. What happened? \end{center}
Where our candidate answers would be: \textit{[PersonX puts out a fire, PersonX puts out a water]}.

For inference concept prediction, we state the event and follow it with ``What happens next?". We then create inference candidates with a ground truth salient concept and an antonym concept.  For example:

\begin{center} PersonX puts out a fire. What happens next? \end{center}
Where our candidate answers would be:  \textit{[PersonX wants to receive recognition, PersonX wants to give recognition]}.

\begin{table*}[th]
    \centering
    \begin{small}
    \begin{tabular}{l|c|cc|ccc|}
    \toprule
    \textbf{} & \textbf{} &\textbf{BERT-FT} & \textbf{} & \textbf{RoBERTa-FT}  & \textbf{} & \textbf{}\\
    \midrule
    \textbf{Type} & \textbf{majority} & \textbf{max} & \textbf{WS} &\textbf{max} & \textbf{WS}\\
    \midrule
    \textbf{Relation Probes} &&\\
    xWant vs xEffect & 56  & 95.25 & 93.56  & 95.45 & 93.95\\
    xWant vs xReact & 52 & 99.10 & 98.70 & 99.21 & 98.32 \\
    xWant vs xIntent & 57  &	75.30 &	66.94 &	70.11 &	59.02 \\
    xWant vs xNeed & 57	 &	82.17	&72.48	&86.70&	70.66\\
    xWant vs xAttr & 52&	99.19	&99.11&	99.22&	98.81 \\
    xEffect vs xReact & 54&	98.17	&96.73&		98.18	&93.32 \\
    xEffect vs xIntent & 51&	95.89	&93.93&		96.30&	93.50\\
    xEffect vs xNeed & 51&	95.22	&93.48&	95.64&	93.86\\
    xEffect vs xAttr & 58&	98.44	&97.58&		98.58&	97.57\\
    xReact vs xIntent &55	&97.96&	97.45&	97.93&	93.76 \\
    xReact vs xNeed &55&	99.27	&98.79&		99.41&	98.60 \\
    xReact vs xAttr &55	&	80.54	&76.48&	79.82	&75.65\\
    xIntent vs xNeed & 50	&87.81&	77.81&	90.00	&72.19\\
    xIntent vs xAttr & 59	&98.42&	98.08&		98.51	&97.78\\
    xNeed vs xAttr &59	&99.14	&98.97	&99.16	&98.54\\
    oWant vs oEffect &61&		93.90	&91.21&	93.98&	90.78\\
    oWant vs oReact &52&		99.10	&98.49&		99.08	&98.81\\
    oEffect vs oReact  & 60&	97.29	&96.26&97.72	&96.55\\
    \midrule
    \textbf{Agent-Patient Probes}&&\\
    xWant vs oWant &70&	84.46&	76.62&	83.70&	72.45\\
    xEffect vs oEffect &75&	84.99	&77.58&		83.45	&76.67\\
    xReact vs oReact & 70&	78.77&	72.49&	70.88	&70.11\\
    \midrule
    \textbf{Concept Probes} &&\\
    inference &-&	94.99&	91.82&	95.26&	87.06\\
    event &-	&98.47&	91.82&		97.41	&91.64 \\
    \bottomrule
    \end{tabular}
    \end{small}
    \caption{Majority label and fine-tuning results for MLM probes.}
    \label{tab:main_results_mlm_probes_detail}
\end{table*}
\subsubsection{MLM}

\paragraph{Relational Probes.}
For inference dimensions relating to PersonX, we map the inference dimensions \texttt{xWant}, \texttt{xNeed}, \texttt{xIntent}, \texttt{xReact}, \texttt{xEffect}, \texttt{xAttr} onto the verbs \textit{wants}, \textit{needs}, \textit{intends}, \textit{feels}, \textit{effect}, and \textit{is}. For example, given the event:

\begin{center}PersonX puts out a fire. \end{center}

We have the following inference in the \texttt{xWant} dimension: \textit{to receive recognition}. We map this onto text for the following probe:

\begin{center}PersonX puts out a fire.  PersonX [MASK] to receive recognition. \end{center}

The correct prediction for [MASK] is \textit{wants}. So for the probe \texttt{xWant} vs  \texttt{xNeed}, our candidate answers for predicting the mask would be \textit{[wants, needs]}.

For inference dimensions relating to Others, we map the inference dimensions \texttt{oWant}, \texttt{oReact}, and \texttt{oEffect} to the same verbs as before:  \textit{wants}, \textit{feels}, and \textit{effect} respectively. We set up probes in the same way as above.

\paragraph{Agent-Patient Probes.}
We create probes to evaluate whether a model can determine whether an inference dimension is assigned to PersonX or others. For example, consider the following probe:

\begin{center}PersonX puts out a fire. [MASK] wants to receive recognition. \end{center}

In this example, the correct prediction is \textit{PersonX}. However, in the below probe, the correct prediction is \textit{others}. 

\begin{center}PersonX puts out a fire. [MASK] want to thank PersonX. \end{center}

Both of these probes use the following answer candidates: \textit{[PersonX, others]}. We also remove plurals to ensure that the model does not make predictions using hints from grammar.

\paragraph{Concept Probes.}
We investigate two kinds of concepts in our probe: event concepts and inference concepts. In event concept probe construction, we find the most salient concept in the event via POS tagging. We then replace this concept with [MASK] and set candidate answers as the ground truth answer and an antonym, as found via WordNet \cite{1995wordnet}. For example, given the event and inference:

\begin{center}PersonX discovers the answer. PersonX feels accomplished. \end{center}

We identify \textit{discovers} as the most salient concept in the event, and use the lemma from WordNet: \textit{discovery}. We then use this to find a viable antonym: \textit{lose}. Finally, we have the probe: 

\begin{center}PersonX [MASK] the answer. PersonX feels accomplished. \end{center}

And the candidates: \textit{[discovery, lose]}.

We lemmatize the answers to allow for fair prediction between the truth concept and the antonym (which often comes lemmatized from Wordnet). 

Similarly, we construct inference concept probes by predicting salient concepts in the inference dimension instead of the event.

\subsubsection{Training Setup}
We create a training and a development set for each of our probes using the ATOMIC train and dev set. We show the sizes of our probe dev sets in Table \ref{tab:probe_size}. The sizes are directly derived from ATOMIC dev sizes. We train each model using 1 GeForce GTX 1080 Ti GPU. 

\subsection{Commonsense MLM Probes: Fine-tuned}
\subsubsection{Setup}
We evaluate our MLM probes in a fine-tuned setting. We train on each probe's respective training set and evaluate the max and WS as in \newcite{talmor2019olmpics}, which defines (1) max as the maximal accuracy on the learning curve and (2) WS as the weighted average of accuracies on the learning curve, where higher weights are assigned to earlier points on the curve. This is to examine how fast the model learns given its encoding before fine-tuning.

\subsubsection{Results}
Table \ref{tab:main_results_mlm_probes_detail}  compares the performance of BERT and RoBERTa for fine-tuned results. 

\noindent\textbf{Fine-tuning Results: }
After fine-tuning on probe training sets, both models do not fully solve the following categories: \texttt{xWant} vs \texttt{xIntent}, \texttt{xWant} vs \texttt{xNeed}, \texttt{xReact} vs \texttt{xAttr}, and \texttt{xIntent} vs \texttt{xNeed}. This demonstrates that these commonsense knowledge categories are difficult to learn even with fine-tuning. Additionally, BERT does learn faster than RoBERTa for the following Subject Probes: \texttt{xWant} vs \texttt{xIntent}, \texttt{xNeed} vs \texttt{xIntent}, and \texttt{xReact} vs \texttt{xIntent}. Overall, this illustrates that BERT and RoBERTa do not capture much ATOMIC commonsense in a zero-shot setting, and that many of these relations are difficult to learn even with fine-tuning, including nuanced relations like \texttt{xWant} vs \texttt{xIntent} and \texttt{xWant} vs \texttt{xNeed}.
Agent-Patient relations seem difficult to learn and do not achieve high final results. Similarly, BERT and RoBERTa perform poorly on Concept Probes in zero-shot setting, however seem to learn these quickly with high final results. Overall, it seems that both models perform comparably for most probes.

\subsection{Reproducibility}
We train each KS model using 2 GeForce GTX 1080 Ti GPUs. Hyperparameter settings are used from previously reported BERT results on each task \cite{sap2019socialiqa, bisk2019piqa, da2019understanding}.

\end{document}